%% file: conference.tex
\documentclass[conference]{IEEEtran}
\IEEEoverridecommandlockouts
\usepackage{cite}
\usepackage{amsmath,amssymb,amsfonts}
\usepackage{algorithmic}
\usepackage{graphicx}
\usepackage{textcomp}
\usepackage{array}
\usepackage{booktabs}
\usepackage{multirow}
\usepackage{stfloats} 
\usepackage{xcolor}
\usepackage{float}
\usepackage{placeins}
\usepackage{hyperref}
\usepackage{booktabs}

\def\BibTeX{{\rm B\kern-.05em{\sc i\kern-.025em b}\kern-.08em
    T\kern-.1667em\lower.7ex\hbox{E}\kern-.125emX}}

\begin{document}

\title{Integrating Locality-Aware Attention \\ with Transformers for General Geometry PDEs
\thanks{This research was supported by the Institute of Information \& Communica- tions Technology Planning \& Evaluation (IITP) grant, funded by the Korea government (MSIT) (No. RS-2019-II190079 (Artificial Intelligence Graduate School Program (Korea University)), No. RS-2024-00436857 (Information Technology Research Center (ITRC)), and No. RS-2024-00457882 (AI Re- search Hub Project)). \\ \textbf{*} Seong-Whan Lee is the corresponding author.}
}

\author{
\IEEEauthorblockN{
\begin{tabular}{cc}
Minsu Koh & Beom-Chul Park \\
\textit{Dept. of Artificial Intelligence} & \textit{Dept. of Artificial Intelligence} \\
\textit{Korea University, Seoul, South Korea} & \textit{Korea University, Seoul, South Korea} \\
minsukoh@korea.ac.kr & sea\_breeze88@korea.ac.kr
\end{tabular}
}
\\ 
\IEEEauthorblockN{
\begin{tabular}{cc}
Heejo Kong & Seong-Whan Lee\textbf{*} \\
\textit{Dept. of Brain and Cognitive Engineering} & \textit{Dept. of Artificial Intelligence} \\
\textit{Korea University, Seoul, South Korea} & \textit{Korea University, Seoul, South Korea} \\
hj\_kong@korea.ac.kr & sw.lee@korea.ac.kr
\end{tabular}
}
}

\maketitle

\begin{abstract}
Neural operators have emerged as promising frameworks for learning mappings governed by partial differential equations (PDEs), serving as data-driven alternatives to traditional numerical methods. While methods such as the Fourier neural operator (FNO) have demonstrated notable performance, their reliance on uniform grids restricts their applicability to complex geometries and irregular meshes. Recently, Transformer-based neural operators with linear attention mechanisms have shown potential in overcoming these limitations for large-scale PDE simulations. However, these approaches predominantly emphasize global feature aggregation, often overlooking fine-scale dynamics and localized PDE behaviors essential for accurate solutions. To address these challenges, we propose the \textit{Locality-Aware Attention Transformer (LA2Former)}, which leverages \(K\)-nearest neighbors for dynamic patchifying and integrates global-local attention for enhanced PDE modeling. By combining linear attention for efficient global context encoding with pairwise attention for capturing intricate local interactions, LA2Former achieves an optimal balance between computational efficiency and predictive accuracy. Extensive evaluations across six benchmark datasets demonstrate that LA2Former improves predictive accuracy by over 50\% relative to existing linear attention methods, while also outperforming full pairwise attention under optimal conditions. This work underscores the critical importance of localized feature learning in advancing Transformer-based neural operators for solving PDEs on complex and irregular domains. Code is available at \href{https://github.com/komingsu/LA2Former}{https://github.com/komingsu/LA2Former}.
\end{abstract}


\section{Introduction}
Partial differential equations (PDEs) underlie a wide range of phenomena in fluid dynamics, materials science, meteorology, and geophysics. Traditional numerical methods, such as the finite element method (FEM) \cite{FEM} and the finite difference method (FDM) \cite{FDM}, have been widely used, but they often incur high computational costs due to large discretizations and repeated iterative solutions. These costs can become prohibitive for real-time decision-making or iterative design tasks, such as weather forecasting and fluid flow analysis \cite{PDE_weather}.

\input{figures/KNN_gather_visual}

Recent advances in deep neural networks have stimulated interest in neural operators that aim to directly learn mappings between input and output functions governed by PDEs \cite{FNO, DeepONet, Galerkin_Transformer, PINN}. Among these approaches, the Fourier neural operator (FNO)~\cite{FNO} has demonstrated remarkable performance by using spectral-domain representations to capture the intricate mappings inherent in PDE solutions. However, its reliance on uniform meshes limits its flexibility and efficiency when applied to irregular meshes or complex geometries \cite{CNO}. Therefore, methods such as GINO \cite{GINO, GNO} and Geo-FNO \cite{Geo-FNO} have been proposed to handle complex geometries. Nevertheless, issues such as increased overhead, loss of fine detail during downsampling, and limited scalability have persisted \cite{U-FNO, U-NO, ReNO, CNO, swl1, swl2, swl3, swl4}.

Transformers \cite{Transformer_ori}, a notable breakthrough in many fields \cite{Bert, ViT, GNOT}, offer an efficient approach to geometry-related problems by representing all data as sequences. However, standard attention exhibits quadratic complexity in the number of mesh points, making large-scale, high-fidelity simulations computationally expensive. This increased complexity not only drives up computational overhead but also complicates the extraction of meaningful correlations among large numbers of mesh points \cite{SwinT, ONO}. To alleviate these concerns, neural operators leveraging linear attention mechanisms have been introduced \cite{Linear_Transformer, Galerkin_Transformer, OFormer, FactFormer}. While linear attention substantially reduces overhead by shifting from pairwise interactions to feature-by-feature summarization, it often fails to capture the local correlations and complex multi-physics interactions critical for accurate PDE solutions \cite{Patchesneed}. Therefore, more efficient approaches capable of handling large meshes while preserving essential physical correlations are still needed.

\input{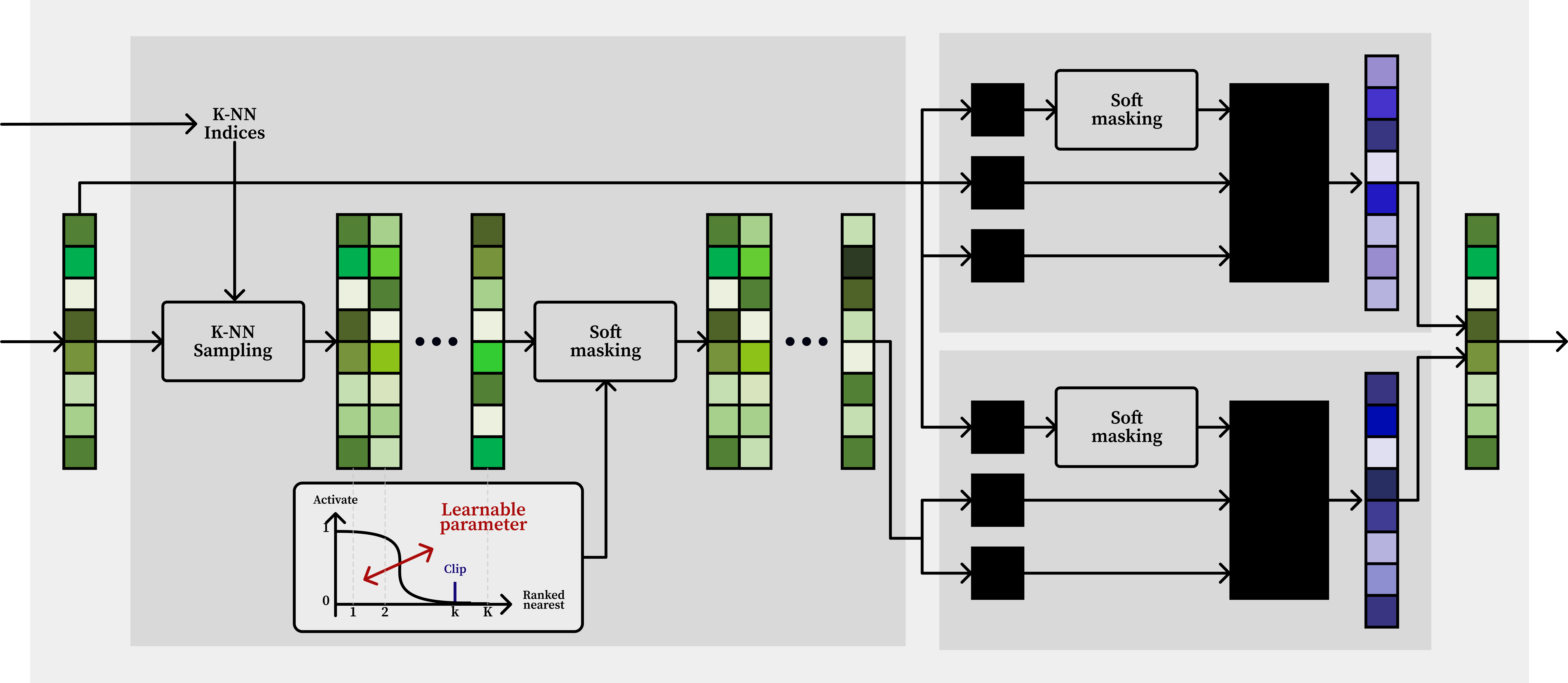}

To overcome these challenges, we propose the Locality-Aware Attention Transformer (LA2Former), a novel framework developed to meet the computational efficiency and accuracy requirements for neural operators. Our method focuses on three core objectives:  
\textit{i)} \textbf{Flexibility}: Provide a unified framework applicable to diverse geometries and mesh configurations. This flexibility ensures that LA2Former can seamlessly adapt to various PDE problems, from simple uniform grids to highly complex and irregular domains.  
\textit{ii)} \textbf{Integral Kernels}: Enable accurate approximation of both global and local physical behaviors governed by PDEs. By learning integral kernels that reflect the underlying physics, LA2Former seeks robust performance across multi-physics interactions.  
\textit{iii)} \textbf{Efficient Attention Mechanism}: Maintain computational efficiency comparable to linear attention while still capturing critical local correlations. Specifically, LA2Former aims to mitigate the quadratic cost barrier of standard attention without sacrificing essential fine-scale details.

As shown in Figure~\ref{fig:illustrationKNN}, the LA2Former employs a Locality-Aware patchifying module, which leverages \(K\)-nearest neighbors to dynamically capture local neighborhood information. This design ensures the effective extraction of meaningful local relationships, enabling robust and context-aware feature representation. Furthermore, the architecture integrates local interactions and global dependencies through a global-local attention mechanism. The global attention component employs linear attention with a computational complexity of \(O(N\times C^2)\), facilitating efficient modeling of global physical information. In contrast, the local attention component adopts standard attention with a computational complexity of \(O(N \times K \times C)\), making it particularly effective in capturing intricate local physical phenomena.

We evaluated the LA2Former across six industry-standard benchmarks~\cite{Geo-FNO, FNO} and observed consistently superior performance. Specifically, datasets related to Darcy flow, elasticity, airfoil dynamics, and plasticity emphasized the model’s state-of-the-art capabilities. Experiments conducted by varying the fixed window size \(K\) revealed its role in enabling a trade-off between predictive performance and computational efficiency. The results demonstrated that an optimal window size achieved lower error rates compared to full pairwise self-attention mechanisms, which exhaustively consider all pairwise data point interactions, while simultaneously reducing computational costs. Moreover, comprehensive ablation studies were performed to quantify the contributions of individual architectural components, further validating the efficacy of the proposed framework.

\section{Related works}

\subsection{Fourier-based Neural Operators}

Neural operators have been actively investigated for PDE-constrained problems, aiming to learn mappings between infinite-dimensional function spaces. The FNO~\cite{FNO} is a representative approach that employs spectral-domain representations to approximate integral operators on uniform grids. Although FNO achieves efficient inference, its reliance on regular meshes and periodic assumptions can degrade performance for complex or irregular domains.

Subsequent efforts have expanded FNO’s scope. For instance, U-FNO~\cite{U-FNO} adopts a hierarchical U-Net-like architecture for multi-scale feature extraction, and geo-FNO~\cite{Geo-FNO} projects unstructured meshes onto a latent regular grid. However, these methods may still encounter performance or scalability issues with highly non-periodic geometries. Other variants like GINO~\cite{GINO} and CNO~\cite{CNO} incorporate geometry encoders or continuous operator perspectives to tackle domain irregularities, yet computational overhead and mesh complexity remain problematic. Overall, Fourier-based methods exhibit a trade-off between capturing global solution characteristics and preserving local high-frequency details, especially on unstructured meshes.

\subsection{Transformer-based Models with Efficient Attention}

Transformers~\cite{Transformer_ori} have shown promise in PDE modeling due to their ability to capture long-range dependencies. However, naive self-attention incurs quadratic complexity in the number of mesh points, making it less feasible for large-scale or high-fidelity simulations.

Various approaches reduce this overhead. Galerkin Transformer~\cite{Galerkin_Transformer} employs a linear attention mechanism \cite{Linear_Transformer} that replaces direct pairwise token interactions with a feature-wise aggregation. This approach projects query and key vectors into a lower-dimensional space, enabling indirect interactions through compact global representations. Consequently, the computational complexity is reduced from quadratic to linear growth with respect to the number of tokens, achieving significant efficiency gains while retaining the ability to capture global dependencies. OFormer~\cite{OFormer} uses a similar linear attention within an encoder–decoder structure, improving scalability on irregular domains.

Meanwhile, hierarchical strategies like HT-Net~\cite{HT_NET} adopt a local patch concept inspired by Swin Transformers~\cite{SwinT}, albeit not identically. This hierarchical design effectively captures localized features, but the reliance on regular grid partitioning can limit its applicability to more complex or irregular meshes. Other approaches, such as FactFormer~\cite{FactFormer} and GNOT~\cite{GNOT}, decompose attention or use domain decomposition to handle multi-scale PDEs. Still, balancing local detail against global coverage remains challenging.

In summary, methods based on the FNO often struggle to handle irregular meshes effectively without losing critical frequency information. Similarly, Transformer-based approaches face a trade-off between capturing global attention and modeling fine-grained local interactions. To address these limitations, we propose LA2Former, which incorporates a KNN-based locality-aware patchifying module combined with a dual-attention design. This hybrid mechanism achieves near-linear computational complexity while preserving essential local correlations, thereby bridging the gap between Fourier-based operators and Transformer-driven PDE solvers.

\section{Preliminary}

\noindent\textbf{Problem Definition.}  
Physical systems governed by PDE often involve mapping an external forcing function, boundary/initial condition, or similar input configuration to the corresponding solution of the PDE. In the operator-learning framework, this problem is formulated as learning a mapping \(\mathcal{G}: \mathcal{F} \to \mathcal{U}\), where \(\mathcal{F}\) and \(\mathcal{U}\) are function spaces (e.g., \(L^2(\Omega)\)). The operator \(\mathcal{G}\) takes an input function \(f \in \mathcal{F}\) and returns the solution \(u \in \mathcal{U}\).
Formally, given a dataset
\begin{equation}
\mathcal{S} = \{(f_i, u_i)\}_{i=1}^N, \quad \text{where } u_i = \mathcal{G}(f_i),
\end{equation}
the objective is to learn a parameterized model \(\mathcal{G}_\theta\) that approximates the true operator \(\mathcal{G}\). This approach has practical utility in scenarios where classical numerical solutions must be repeatedly computed for varied inputs, as learning \(\mathcal{G}\) can substantially reduce the overall computational cost.

\vspace{0.1cm}
\noindent\textbf{Discretization for Practical Implementation.}  
Implementing \(\mathcal{G}\) directly on continuous functions defined over \(\Omega \subset \mathbb{R}^{C_s}\) can be prohibitively complex. Instead, the domain \(\Omega\) is discretized into a mesh
\begin{equation}
\mathbf{X} = \{x_j \in \Omega\}_{j=1}^M,
\end{equation}
where \(M\) is selected based on factors such as desired spatial resolution. Each input function \(f_i\) is sampled at these mesh points to form a matrix \(f_i(x_j)\in\mathbb{R}^{M\times C_f}\), and similarly, the output function is recorded as \(u_i(x_j)\in\mathbb{R}^{M\times C_u}\). This strategy reduces the infinite-dimensional operator-learning problem to learning a finite-dimensional mapping between high-dimensional spaces. Methods such as spectral approximations, finite element methods, or finite difference schemes may be used for this discretization. By working with discretized representations, neural operators can be trained using standard deep learning frameworks.

\vspace{0.1cm}
\noindent\textbf{Data Fitting via Loss Minimization.}  
To train \(\mathcal{G}_\theta\), we minimize the discrepancy between \(\mathcal{G}_\theta(f_i)\) and the reference solutions \(u_i\). A common choice is the relative \(L^2\)-norm,
\begin{equation}
\mathcal{L}(\theta) = \frac{1}{N} \sum_{i=1}^N 
\frac{\|\mathcal{G}_\theta(f_i) - u_i\|_2^2}{\|u_i\|_2^2},
\end{equation}
where \(\|\cdot\|_2\) denotes the discrete \(L^2\)-norm. Normalizing by \(\|u_i\|_2^2\) helps mitigate scale differences across diverse PDE problems. Iterative updates to \(\theta\) with this loss enable the neural operator to learn and generalize the underlying mapping.

\section{Methodology}
\subsection{Overview}
The LA2Former network architecture consists of \(L\) layers (\(\mathcal{K}^{(l)}\)) operating on the encoded input produced by an encoder \(\mathcal{E}\), with the final prediction represented by \(\mathcal{P}\), as shown in Eq.~\eqref{eq4}:  
\begin{equation}  
G_{\theta} = \mathcal{P} \circ \mathcal{K}^{(L)} \circ \mathcal{K}^{(L-1)} \circ \cdots \circ \mathcal{K}^{(1)} \circ \mathcal{E},  
\label{eq4}  
\end{equation}  
where \(\mathcal{E}\) employs multilayer perceptrons (MLPs) to transform the input \(f_i \in \mathbb{R}^{M \times C_f}\) into a hidden state \(h_i^{(1)} \in \mathbb{R}^{M \times C}\).

Each layer \(\mathcal{K}^{(l)}\) introduces a learnable kernel operator for integral computations as defined in Eq.~\eqref{eq5}:  
\begin{equation}  
h^{(l)}(x) = \int_{\Omega} \kappa_l(x, y)\, h^{(l-1)}(y)\, dy,  
\label{eq5}  
\end{equation}  
where \(\kappa_l(x, y)\) represents the learnable kernel function at layer \(l\), and \(\Omega\) is the physical domain. The final output \(h^{(L)} \in \mathbb{R}^{M \times C}\) is projected through a linear module \(\mathcal{P}\), yielding the PDE solution \(u \in \mathbb{R}^{M \times C_u}\).  

Inspired by the Transformer architecture \cite{Transformer_ori}, each layer combines global and local attention through the proposed global-local attention (GLA) module. This layer follows a two-stage block as shown in Eq.~\eqref{eq6}:  
\begin{equation}  
\begin{split}  
\hat{h}^{(l)} &= \text{GLA}(\text{LayerNorm}(h^{(l-1)})) + h^{(l-1)}, \\  
h^{(l)} &= \text{FeedForward}(\text{LayerNorm}(\hat{h}^{(l)})) + \hat{h}^{(l)},  
\end{split}  
\label{eq6}  
\end{equation}  
where the GLA module integrates linear attention for global feature extraction and local attention for neighborhood interactions.

\subsection{Locality-Aware Encoding}

\noindent\textbf{Instant KNN Patches.}  
The proposed approach encodes local neighborhood information across general geometric domains by employing the K-nearest neighbors (KNN) method with a dynamically controlled soft mask. For a discretized domain \( X \in \mathbb{R}^{M \times C_s} \), pairwise distances between points are computed to construct a distance matrix \(\mathbf{D} \in \mathbb{R}^{M \times M}\), where each entry represents the Euclidean distance between two points in \( X \). From this distance matrix, the indices of the \( K \)-nearest neighbors for each point are determined, yielding the KNN index matrix \(\mathbf{I}_{\text{KNN}} \in \mathbb{R}^{M \times K}\). 

The KNN index matrix \(\mathbf{I}_{\text{KNN}}\) is utilized to aggregate features from the normalized hidden state \(\bar{h} = \text{LayerNorm}(h^{(l)}) \in \mathbb{R}^{M \times C}\) at the \( l \)-th layer. The KNN feature representation \(\bar{h}_{\text{KNN}}^{(l)}\) is defined as:
\begin{equation}
\bar{h}_{\text{KNN}}^{(l)}[a, b, :] = \bar{h}[\mathbf{I}[a, b], :], 
\quad 
\begin{aligned}
&\forall a \in \{1, \dots, M\} \\
&\forall b \in \{1, \dots, K\}
\end{aligned}
\label{eq:knn_feature_representation}
\end{equation}
Here, \(\bar{h}_{\text{KNN}}^{(l)} \in \mathbb{R}^{M \times K \times C}\) encapsulates the local neighborhood information of \(\bar{h}\), preserving the geometric relationships of \( X \).

\vspace{0.1cm}
\noindent\textbf{Dynamic Soft Mask for Adaptive Selection.}  
To adaptively control the number of neighbors, a soft weighting mask is applied to the gathered features. The weights \( \mathbf{W}_K = \{w_k\}_{k=1}^K \) are computed as:
\begin{equation}
w_k = \sigma\left(-\alpha \left(k - \sigma(s) \cdot (K - 1) - 1 \right)\right),
\label{eq:soft_weighting_mask}
\end{equation} 
where \( \mathbf{W}_K \in \mathbb{R}^K \) represents the weights associated with the \( K \)-nearest neighbors. Here, \( \sigma \) denotes the sigmoid function, \( \alpha \) regulates the sharpness of the weighting curve, and \( s \) is a learnable parameter determining the effective number of neighbors.

The dynamically weighted KNN feature representation is obtained through element-wise multiplication of \(\bar{h}_{\text{KNN}}^{(l)}\) with the soft weighting mask \( \mathbf{W}_K \), resulting in:
\begin{equation}
\tilde{h}_{\text{KNN}}^{(l)} = \bar{h}_{\text{KNN}}^{(l)} \odot \mathbf{W}_K,
\label{eq:final_knn_representation}
\end{equation}
where \(\tilde{h}_{\text{KNN}}^{(l)} \in \mathbb{R}^{M \times K \times C}\) encodes both geometric locality and adaptive feature weighting.

\subsection{Global-Local Attention (GLA)}

The global-local attention mechanism is designed to capture both long-range dependencies and local interactions within complex multidimensional data. It integrates global attention for contextual relationships and local attention for neighborhood-level details, providing a comprehensive model of the underlying structures:
\begin{equation}
\label{eq:gla_combined}
\mathrm{GLA}\bigl(\bar{h}\bigr) = \mathrm{Linear}\Bigl(
\mathrm{Concat}\bigl(
\mathrm{G}(\bar{h}),\,
\mathrm{L}(\bar{h}, \tilde{h}_{\text{KNN}}^{(l)})
\bigr)
\Bigr),
\end{equation}
where \( \bar{h} = \mathrm{LayerNorm}(h^{(l-1)}) \in \mathbb{R}^{M \times C} \) and \(\tilde{h}_{\text{KNN}}^{(l)}\) is the neighborhood-aware representation defined in Equation \eqref{eq:final_knn_representation}. The terms \(\mathrm{G}(\cdot)\) and \(\mathrm{L}(\cdot, \cdot)\) denote the global and local attention mechanisms, respectively. The output \(\mathrm{GLA}(\cdot)\) resides in \(\mathbb{R}^{M \times C}\).

\vspace{0.1cm}
\noindent\textbf{Global Attention.}  
Global attention, denoted as \(\mathrm{G}(\cdot)\), captures relationships spanning the entire data point set (e.g., grid points or mesh nodes) by leveraging a linear attention mechanism. The layer-normalized input \(\bar{h} \in \mathbb{R}^{M \times C}\) is linearly projected into query \(Q_g\), key \(K_g\), and value \(V_g\) matrices:
\begin{equation}
Q_g, \; K_g, \; V_g = \mathrm{Linear}(\bar{h}), \quad Q_g, \; K_g, \; V_g \in \mathbb{R}^{M \times d}.
\end{equation}
Normalized queries and keys are derived using the \(\ell_1\)-norm:
\begin{equation}
\|Q_g\|_1 = \sum_{i=1}^{d} |Q_{g,i}|, \quad \|K_g\|_1 = \sum_{i=1}^{d} |K_{g,i}|,
\end{equation}
yielding \(\tilde{Q}_g = Q_g / \|Q_g\|_1\) and \(\tilde{K}_g = K_g / \|K_g\|_1\). These normalized matrices enable efficient computation of global interactions, with the final global attention output given as:
\begin{equation}
\mathrm{G}(\bar{h}) = \tilde{Q}_g \, (\tilde{K}_g^\top V_g) \, D_g^{-1} + \tilde{Q}_g \in \mathbb{R}^{M \times d},
\end{equation}
where \(D_g^{-1}\) serves as an inverse scaling factor derived from the cumulative interaction terms.

This global attention mechanism is particularly effective for modeling long-range dependencies, which are critical for understanding global patterns and maintaining the overall structural integrity of the data. By normalizing queries and keys, the approach ensures numerical stability and scalability, even in high-dimensional settings.

\vspace{0.1cm}
\noindent\textbf{Local Attention.}  
Local attention, denoted as \(\mathrm{L}(\cdot,\cdot)\), focuses on fine-grained interactions between each data point and its dynamically weighted neighborhood. The representation \(\tilde{h}_{\text{KNN}}^{(l)}\) is linearly projected into keys \(K_l\) and values \(V_l\), while \(\bar{h}\) is projected into query \(Q_l\):
\begin{equation}
Q_l = \mathrm{Linear}(\bar{h}), \quad K_l, \; V_l = \mathrm{Linear}(\tilde{h}_{\text{KNN}}^{(l)}).
\end{equation}
The attention scores are computed as:
\begin{equation}
\mathrm{L}(\bar{h}, \tilde{h}_{\text{KNN}}^{(l)}) = \mathrm{Softmax}\left(\frac{Q_l K_l^\top}{\sqrt{d}}\right) V_l.
\end{equation}

This local attention mechanism enhances the model’s ability to capture detailed spatial relationships and subtle variations within the immediate neighborhood of each data point. Such detailed modeling is essential for resolving localized phenomena and preserving fine-grained features, which are often overlooked in global attention mechanisms.

The integration of global and local attention within the GLA module ensures that the model achieves a robust balance between global coherence and local precision. By combining contextual awareness with detailed local interactions, the GLA mechanism effectively supports complex multidimensional data analysis.

\input{tables/benchmark_detail.tex}
\input{tables/main_results}

\section{Experiments}

We conducted experiments across diverse geometries and tasks to demonstrate the effectiveness of incorporating local information into the learning process. The proposed method was also evaluated against state-of-the-art (SOTA) models on widely recognized benchmarks, thereby showcasing its superior performance in various settings.

\subsection{Benchmarks}

\noindent\textbf{Datasets.} Our experiments span a diverse set of PDE tasks (Elasticity, Plasticity, Airfoil, Pipe, Navier–Stokes, Darcy) and geometric configurations (point clouds, structured meshes, and regular grids) in both 2D and (2+1)D settings. As shown in Table~\ref{tab:benchmark_detail}, each benchmark differs in the number of mesh elements (\#Mesh), spatial dimensionality (\#Dim), input-output features, and the total dataset size (\#Dataset). For example, Elasticity relies on 972 point-cloud elements to predict stress distributions in elastic materials, whereas Darcy uses an \(85 \times 85\) regular grid to model fluid flow through porous media. These datasets, introduced by FNO \cite{FNO} and geo-FNO \cite{Geo-FNO}, collectively represent standard benchmarks in neural operator research, enabling a thorough evaluation of model performance on both simple and complex domain geometries.

\vspace{0.1cm}
\noindent\textbf{Baselines.} We evaluate our model against a comprehensive set of baselines, categorized into Fourier-based models and Transformer-based models employing efficient attention mechanisms. Our comparisons include the foundational FNO \cite{FNO} and its advanced variants, such as Geo-FNO \cite{Geo-FNO}, F-FNO \cite{FFNO}, U-FNO \cite{U-FNO} and \cite{LSM}. We also benchmark against several Transformer-based PDE solvers that employ efficient attention mechanisms. These include the Galerkin Transformer \cite{Galerkin_Transformer}, HT-Net \cite{HT_NET}, OFormer \cite{OFormer}, GNOT \cite{GNOT} and FactFormer \cite{FactFormer}.

\vspace{0.1cm}
\noindent\textbf{Implementation Details.} To ensure fairness, the number of layers \( L \) is set to 8, and the hidden feature dimension \( C \) is fixed at 128. For both the global and local attention modules, the feature dimensions are halved to 64. The model uses the relative \( L_2 \) error as both the training loss and the evaluation metric. Experiments on the Elasticity, Plasticity, Airfoil, and Darcy datasets were conducted using an NVIDIA RTX 3090 GPU, while the Pipe and Navier–Stokes datasets were evaluated on an NVIDIA RTX 6000 GPU. All experiments were performed on a single GPU and repeated three times to ensure robustness.

\input{figures/Main_Results_Fig}

\subsection{Results}

\noindent\textbf{Standard Benchmarks.} To comprehensively evaluate our neural operator, we tested the six aforementioned benchmark datasets, encompassing both solid mechanics (Elasticity, Plasticity) and fluid dynamics (Airfoil, Pipe, Navier-Stokes, Darcy). Our proposed model, LA2Former, achieves SOTA performance on the Elasticity, Plasticity, Airfoil, and Darcy datasets, demonstrating its adaptability across structured and unstructured geometries.

By integrating the linear attention mechanism from the Galerkin Transformer as a global module with our proposed local attention mechanism, LA2Former effectively captures fine-grained physical interactions, yielding significant error reductions of 77.5\% on Elasticity, 86.7\% on Plasticity, 52.5\% on Airfoil, and 46.4\% on Darcy. These results emphasize the model's ability to handle point clouds and irregular meshes, where local interactions are critical for accurate physical modeling.

\input{figures/window_size}


Despite these strengths, the hybrid attention framework exhibits limitations on the Navier-Stokes dataset, which uses uniform grids and symmetric boundary conditions. In this setting, the global flow patterns remain relatively consistent, reducing the relevance of localized interactions. Additionally, since the domain lacks obstacles, distinct local features are less pronounced, limiting the effectiveness of local attention and in some cases leading to redundant computations. Nevertheless, in most real-world simulation scenarios, where geometries and boundary conditions play a more significant role, the proposed approach is highly effective at capturing complex physical interactions.

\vspace{0.1cm}
\noindent\textbf{Impact of Local Attention Window Size.} Figure~\ref{fig:windowsize} shows how varying the local attention window size \(K\) affects both the relative \(L_2\) error and the computational cost for the Elasticity and Darcy datasets. In LA2Former, global attention delivers broad context, while local attention targets fine-scale physics. Balancing these two mechanisms is central to achieving accurate and efficient PDE solutions.

For Elasticity, which involves 972 mesh points in a relatively small, sparse domain, a window size of around \(K = 30\) effectively captures localized deformations. Larger values (e.g., \(K > 70\)) introduce unnecessary context that can dilute the focus on local details, causing performance to decline. Meanwhile, the Darcy dataset comprises 7225 mesh points and features multi-scale complexity, making a larger local window (e.g., \(K \approx 40\)) advantageous. Notably, LA2Former outperforms purely global-attention baselines within certain ranges of \(K\), underscoring that well-chosen local regions can outperform naive full-context approaches.

As \(K\) increases, runtime also grows, with denser meshes such as Darcy experiencing steeper computational overhead. Consequently, practitioners should tune \(K\) based on domain complexity and mesh resolution: smaller or simpler domains benefit from compact local windows, whereas larger or more complex domains profit from more extensive ones. These findings confirm the robustness of a hybrid local–global attention strategy in addressing a wide range of PDE problems.

\input{figures/scale.tex}

\vspace{0.1cm}
\noindent\textbf{Ablation Study on Architecture Scaling.}  Figure~\ref{fig:scale_hori} also examines how scaling the model’s \textit{Width} (hidden state size) and \textit{Depth} (number of layers) influences the relative \(L_2\) error for both Elasticity and Darcy. Increasing the hidden dimension boosts feature expressiveness up to roughly 96 channels, beyond which returns begin to diminish. This plateau suggests that an overly large hidden dimension merely adds computational costs without commensurate accuracy gains.

Scaling the depth leads to a more gradual improvement. Elasticity, with fewer mesh points and simpler physics, benefits significantly up to about six layers; performance improvements taper thereafter. Darcy, on the other hand, is multi-scale and continues to show gains up to around 12 layers, although the returns beyond that point are minor. These results suggest that users should choose moderate width and depth for simpler tasks like Elasticity, while more complex problems like Darcy benefit from deeper architectures with at least 96 channels.

\subsection{Discussion}

Our experimental results demonstrate that integrating locality-aware encoding with global-local attention is highly effective in solving a broad spectrum of PDE-based tasks. In particular, LA2Former departs from conventional fixed-window approaches by dynamically adjusting local receptive fields. This strategy allows the model to preserve fine-grained local details while mitigating underfitting in large-scale structural representations, thereby enhancing adaptability across diverse problem domains.

\input{figures/depth_dynamic_k}

A well-defined hierarchical differentiation is observed in the layer representations. Shallow layers primarily capture fine-grained boundary details, whereas deeper layers progressively extend the global context. This hierarchical feature aggregation enhances the balance between high-resolution precision and large-scale structural coherence. As shown in Figure~\ref{fig:dynamic_k}, the evolution of layer-wise attention parameters exhibits distinct patterns in shallow (blue) and deep (red) layers, highlighting the need for flexible receptive field adaptation.

Unlike static local attention mechanisms, LA2Former dynamically adjusts \( K \) at each layer, promoting a more context-aware feature extraction process. Shallow layers adopt smaller \( K \) values to emphasize boundary conditions and localized perturbations, whereas deeper layers employ broader receptive fields to integrate fine-scale details into more comprehensive representations. By adaptively tuning local receptive fields, LA2Former effectively balances feature extraction at different scales, demonstrating superior generalization across PDE domains.

\section{Conclusion}

This paper addresses the challenge of performance degradation in traditional neural operators when essential local correlations in complex physical systems are inadequately captured. To overcome this limitation, we introduced a dynamic, domain-independent patching method that adaptively embeds neighborhood information. By enabling localized attention between adjacent points, this approach significantly enhances modeling accuracy while maintaining computational efficiency.

Despite these advancements, certain challenges persist. While dynamic windowing reduces sensitivity to the choice of maximum window size \( K \), identifying the optimal \( K \) for a given setting remains an open challenge. Additionally, although performing attention on \( K \) local data points for all \( N \) points in the domain is computationally more efficient than full pairwise attention, it still imposes a considerable computational burden. Addressing these challenges will require further optimization, particularly in scaling the approach to extremely large PDE domains.

Extensive experimental evaluations confirm the superiority of the proposed method over existing baselines, demonstrating robust generalization across diverse geometric configurations and irregular datasets. Future research will investigate the integration of a learnable weighting mechanism for dynamically balancing global and local information, thereby mitigating potential asymmetries in attention distribution. Furthermore, hybrid approaches that integrate adaptive local patching with spectral operators hold promise for further improving both efficiency and scalability in large-scale PDE solvers.

\bibliographystyle{ieeetr} 
\bibliography{ref} 

\end{document}

%% file: figures/KNN_gather_visual.tex
\begin{figure}[t]
    \centering
\includegraphics[width=0.95\linewidth]{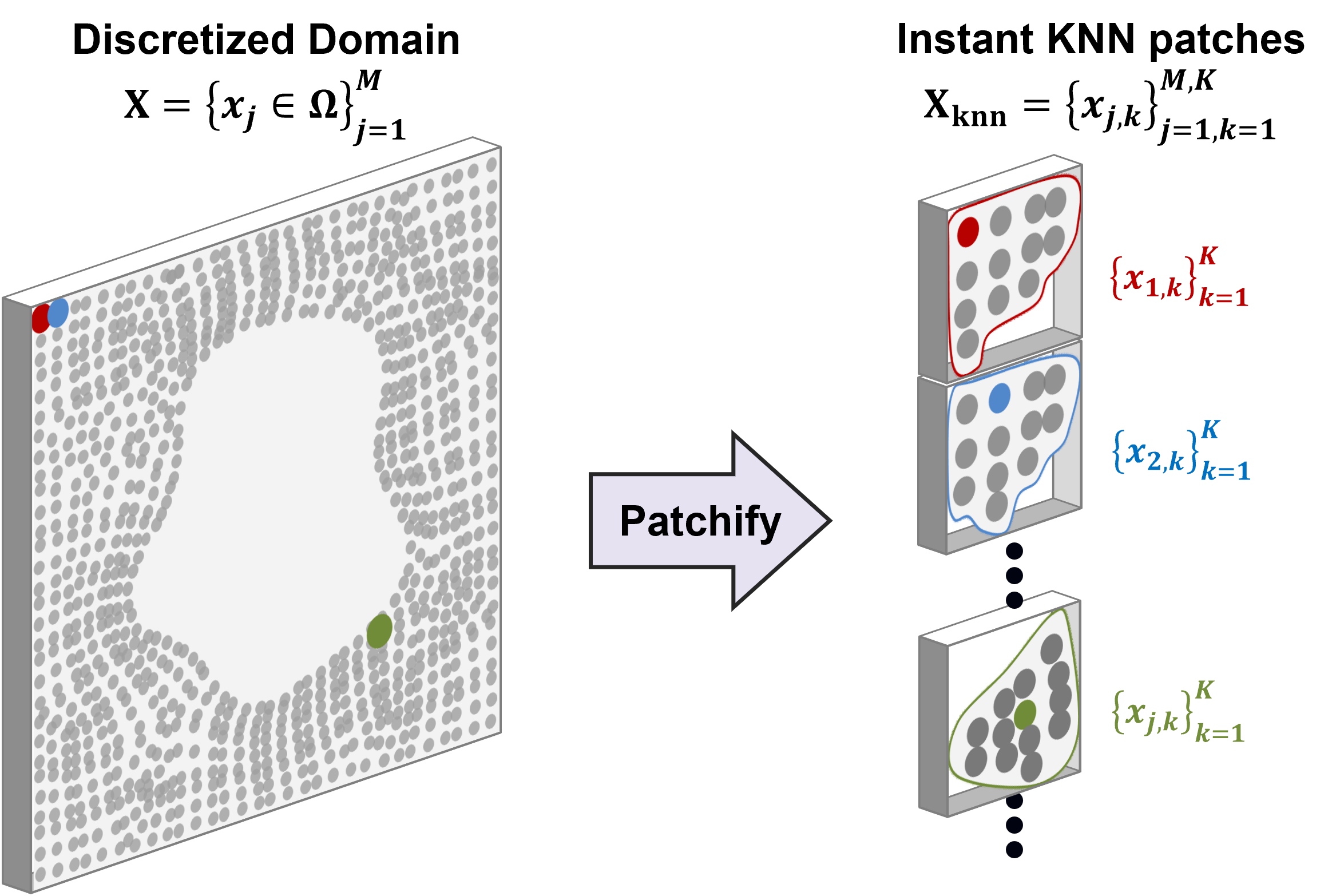}
\caption{
Conceptual illustration of the instant KNN patchifying process proposed in this study. In a discretized 2D domain (left), each point extracts \( k \)-nearest neighbors based on distance to form a \textbf{patch}. The figure visualizes the neighbor sets for selected red, blue, and green points as examples. Note that in this context, a \textbf{patch} refers not to a contiguous spatial division, as in an image, but rather to a set of neighbors defined by distance. The resulting tensor serves as a key input for learning local neighborhood information in subsequent stages.
}
\vspace{-0.4cm}
\label{fig:illustrationKNN}
\end{figure}

%% file: figures/main.tex
\begin{figure*}[t] 
\centering
\includegraphics[width=\linewidth]{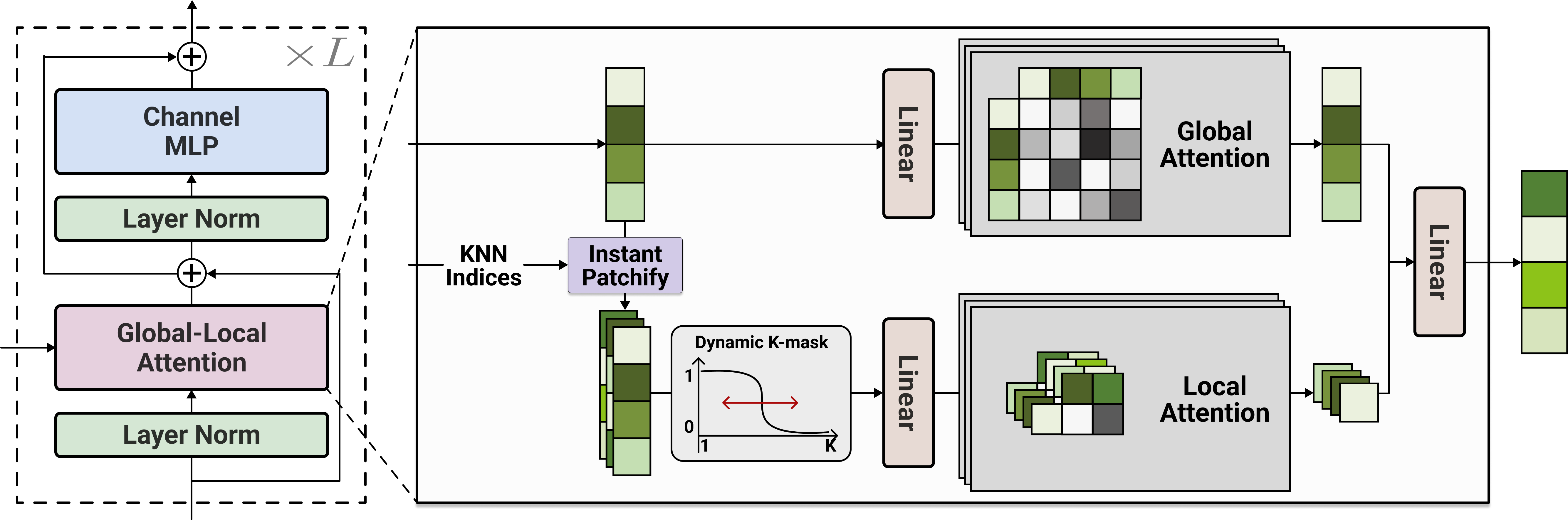} 
\caption{
Schematic overview of the proposed LA2Former layer. The architecture introduces a global-local attention module, which combines global and local interactions to achieve efficient and accurate PDE modeling. At each layer, the discretized input domain is dynamically divided into K-nearest neighbor patches, facilitating parallel computation of global and local attention mechanisms. The outputs from both attentions are integrated to capture long-range dependencies and fine-grained local dynamics, ensuring robust feature representation.
}
\vspace{-0.2cm}
\label{fig:main}
\end{figure*}

%% file: tables/benchmark_detail.tex
\begin{table*}[t]
\caption{Benchmark setup. 
This table summarizes the geometry type, spatial dimensionality (\#Dim), number of mesh elements (\#Mesh), input features (\#Input), tensor shapes (\#Input Tensor, \#Output Tensor), and dataset size (\#Dataset) for each benchmark. A “/” symbol indicates the batch dimension, omitted for simplicity. The input and output tensors follow the structure (Batch, Data Points, Function Dimension, Time Dimension). If the time dimension is absent, only a single time-step is considered.}
\centering
\resizebox{2.0\columnwidth}{!}{
{\renewcommand{\rmdefault}{ptm} 
\renewcommand{\arraystretch}{1.5}
\begin{tabular}{@{}l|c|c|c|c|c|c|c@{}}
\toprule
\textbf{Geometry} & \textbf{Benchmarks} & \textbf{\#Mesh} & \textbf{\#Input} & \textbf{\#Input Tensor} & \textbf{\#Output} & \textbf{\#Output Tensor} & \textbf{\#Dataset} \\
\midrule
Point Cloud & Elasticity & 972 & Structure & (/, 972, 2) & Inner Stress & (/, 972, 1) & (1000, 200) \\
\midrule
\multirow{3}{*}{Structured Mesh} 
  & Plasticity & 101$\times$31 & External Force & (/, 3131, 3) & Mesh Displacement & (/, 3131, 4, 20) & (900, 80) \\
  & Airfoil         & 221$\times$51 & Structure  & (/, 11271, 2)    & Mach Number  & (/, 11271, 1)       & (1000, 200) \\
  & Pipe       & 129$\times$129 & Structure  & (/, 16641, 2)   & Fluid Velocity  & (/, 16641, 1)  & (1000, 200) \\
\midrule
\multirow{2}{*}{Regular Grid}
  & Navier--Stokes & 64$\times$64 & Past Velocity & (/, 4096, 12, 10)  & Future Velocity & (/, 4096, 1, 10) & (1000, 200) \\
  & Darcy          & 85$\times$85 & Porous Medium & (/, 7225, 2) & Fluid Pressure & (/, 7225, 1) & (1000, 200) \\
\bottomrule
\end{tabular}
}}
\label{tab:benchmark_detail}
\end{table*}

%% file: tables/main_results.tex
\begin{table*}[htbp]
\caption{Performance comparison on standard benchmarks.
The table reports the relative $L_2$ error, where lower values indicate better performance. The best-performing results are shown in bold, while the second-best are underlined. A “/” symbol denotes cases where the baseline method is not applicable.}
\centering
{\renewcommand{\rmdefault}{ptm} 
\begin{tabular}{@{}l|c|ccc|cc@{}}
\toprule
& Point Cloud & \multicolumn{3}{c}{Structured Mesh} & \multicolumn{2}{|c}{Regular Grid} \\
\cmidrule{2-2} \cmidrule{3-5} \cmidrule{6-7}
\hspace{2mm}Model & Elasticity & Plasticity & Airfoil & Pipe & Navier-Stokes & Darcy\hspace{2mm} \\
\midrule
\hspace{2mm}FNO \cite{FNO, Geo-FNO} & 0.0229 & 0.0074 & 0.0138 & 0.0067 & 0.1556 & 0.0108 \\
\hspace{2mm}WMT \cite{WMT} & 0.0359 & 0.0076 & 0.0075 & 0.0077 & 0.1541 & 0.0082 \\
\hspace{2mm}U-FNO \cite{U-FNO} & 0.0239 & 0.0039 & 0.0269 & 0.0056 & 0.2231 & 0.0183 \\
\hspace{2mm}F-FNO \cite{FFNO} & 0.0263 & 0.0047 & 0.0078 & 0.0070 & 0.2322 & 0.0077 \\
\hspace{2mm}LSM \cite{LSM} & 0.0225 & 0.0035 & 0.0062 & 0.0050 & 0.1693 & \underline{0.0069} \\
\midrule
\hspace{2mm}Galerkin \cite{Galerkin_Transformer} & 0.0240 & 0.0120 & 0.0118 & 0.0098 & 0.1401 & 0.0084 \\
\hspace{2mm}OFormer \cite{OFormer} & 0.0183 & \underline{0.0017} & 0.0183 & 0.0168 & 0.1705 & 0.0124 \\
\hspace{2mm}GNOT \cite{GNOT} & \underline{0.0086} & 0.0336 & 0.0076 & \textbf{0.0047} & 0.1380 & 0.0105 \\
\hspace{2mm}FactFormer \cite{FactFormer} & / & 0.0312 & 0.0071 & 0.0060 & \underline{0.1214} & 0.0109 \\
\hspace{2mm}ONO \cite{ONO} & 0.0118 & 0.0048 & \underline{0.0061} & 0.0052 & \textbf{0.1195} & 0.0076 \\
\midrule
\textbf{LA2Former (Ours)} & \textbf{0.0054} & \textbf{0.0016} & \textbf{0.0056} & \underline{0.0051} & 0.1274 & \textbf{0.0044} \\
\bottomrule
\end{tabular}
}
\label{tab:performance_main}
\end{table*}

%% file: figures/Main_Results_Fig.tex
\begin{figure}[t]
\begin{center}
\includegraphics[width=1.0\linewidth]{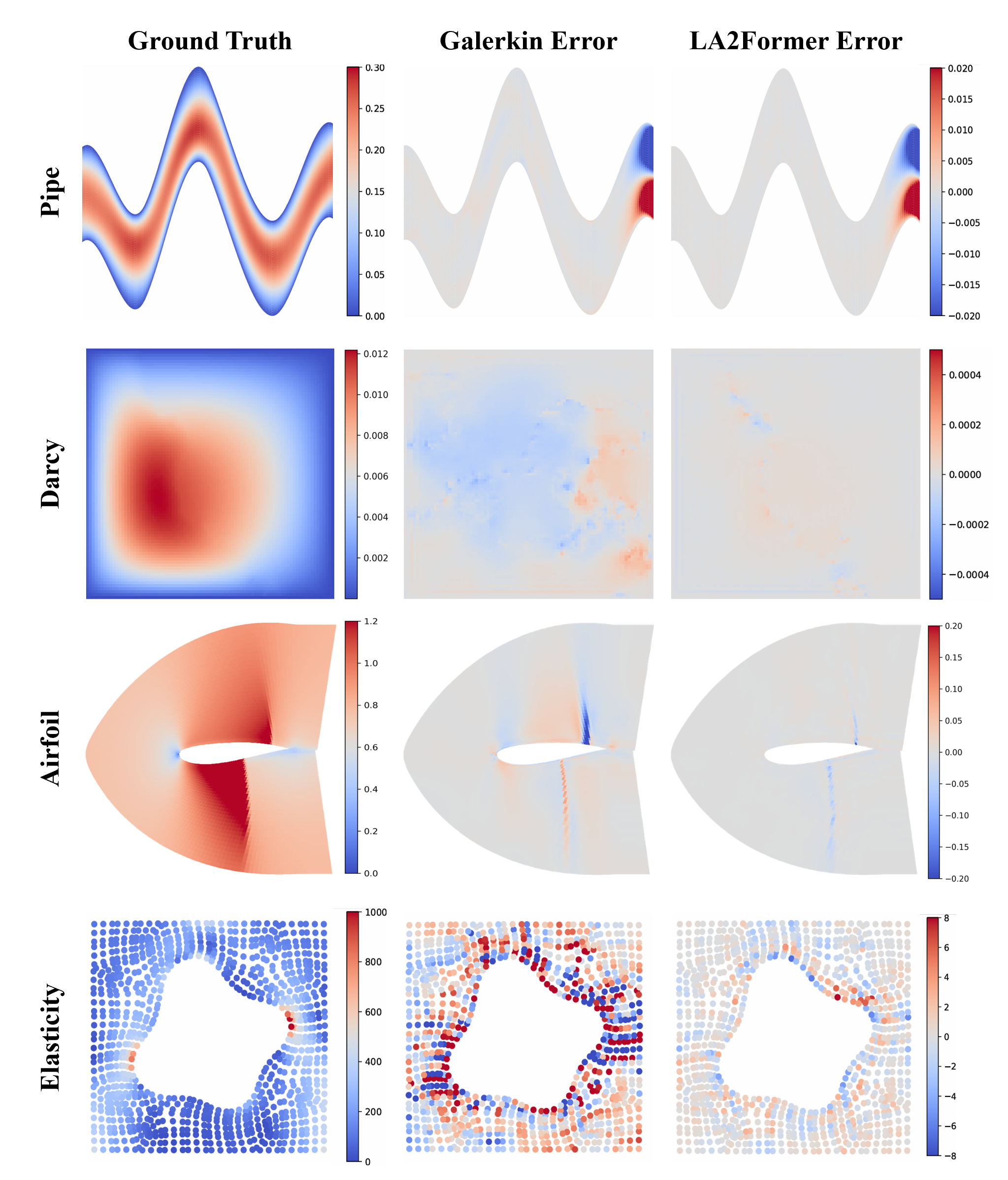}
\end{center}
\vspace{-0.1cm}
\caption{Representative comparisons of the proposed LA2Former versus Galerkin across four PDE tasks. 
Each row shows the ground truth (left), the Galerkin error (center), and the LA2Former error (right).}
\vspace{-0.2cm}
\label{fig:Main_result_fig}
\end{figure}

%% file: figures/window_size.tex
\begin{figure}[t]
\centering
\begin{tabular}{c@{\ }c}
\multicolumn{2}{l}{\textbf{(a) Elasticity}}\\
\includegraphics[height=0.18\textheight]{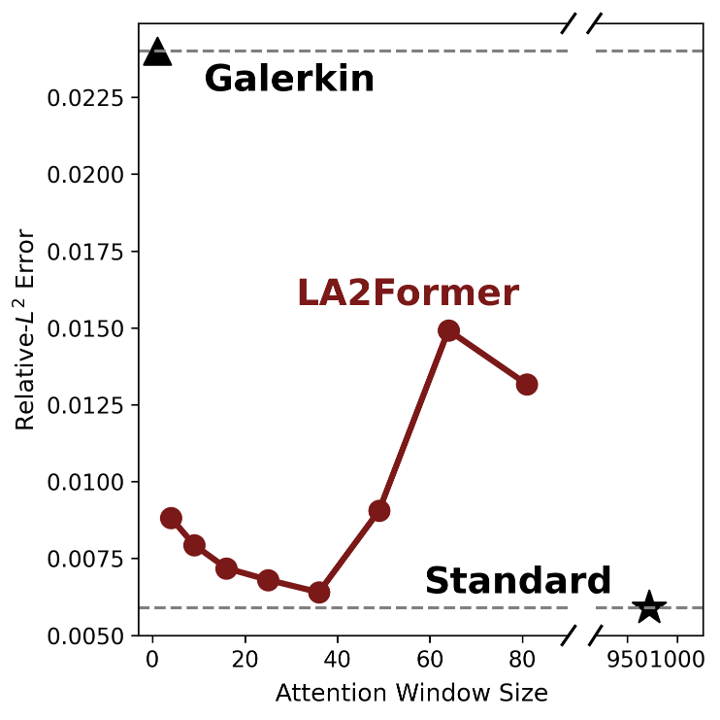} &
\includegraphics[height=0.18\textheight]{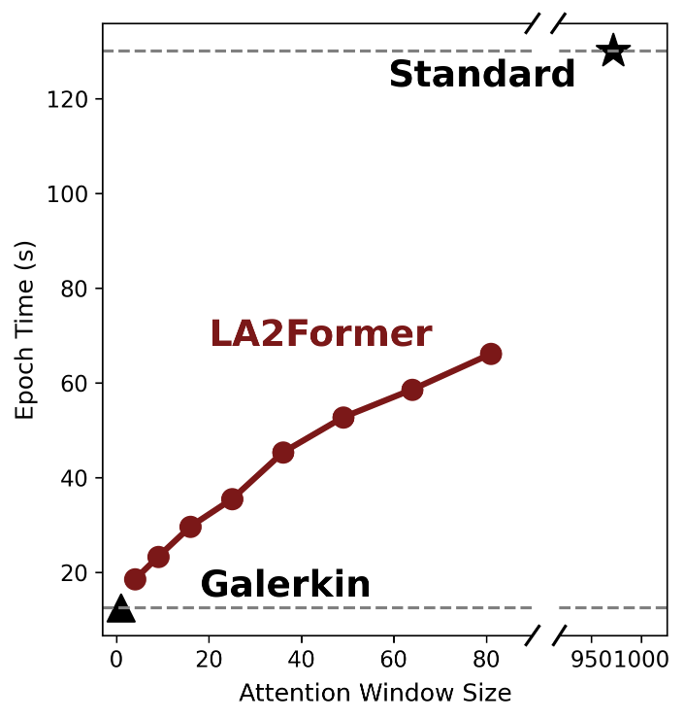} \\
\multicolumn{2}{l}{\textbf{(b) Darcy}}\\
\includegraphics[height=0.18\textheight]{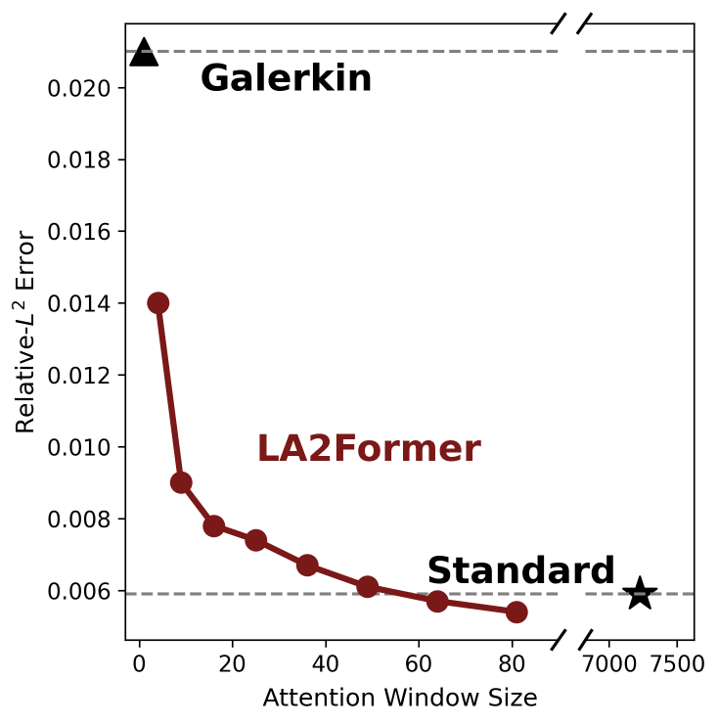} &
\includegraphics[height=0.18\textheight]{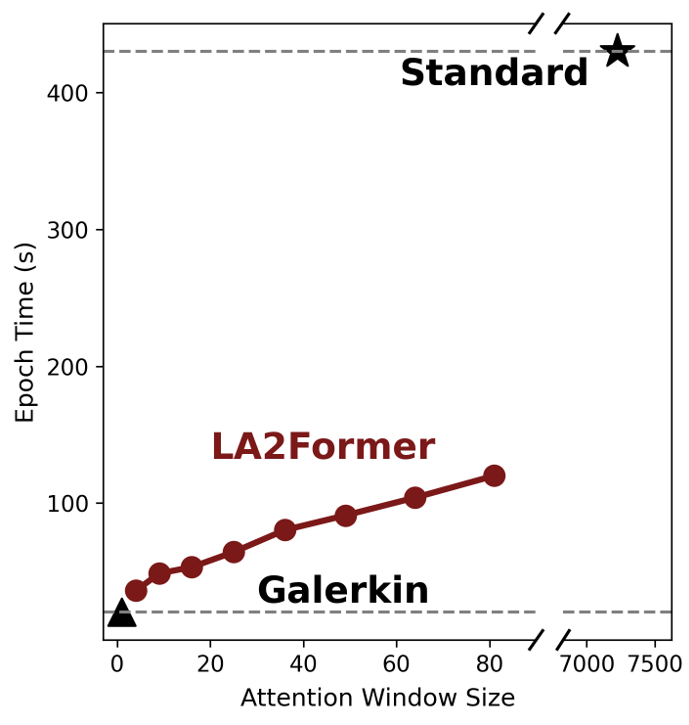} \\
\end{tabular}
\vspace{-0.2cm}
\caption{Comparison of relative $L_2$ error (left) and epoch time (right) with respect to attention window size.
The top row represents the Elasticity dataset, while the bottom row shows the Darcy dataset. In each plot, LA2Former (red circles), Galerkin (black triangles), and Standard (black stars) are contrasted for clarity.}
\label{fig:windowsize}
\vspace{-0.2cm}
\end{figure}

%% file: figures/scale.tex
\begin{figure}[t]
\begin{center}
\includegraphics[width=0.95\columnwidth]{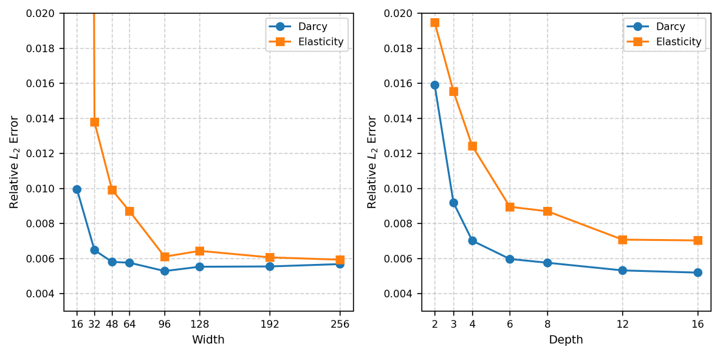} 
\end{center}
\vspace{-0.5cm}
\caption{Effect of width and depth scaling. Relative \( L_2 \) error is shown for varying hidden state sizes (\textit{Width}, left) and number of layers (\textit{Depth}, right) on Darcy and Elasticity datasets.}
\label{fig:scale_hori}
\end{figure}

%% file: figures/depth_dynamic_k.tex
\begin{figure}[t]
\begin{center}
\includegraphics[width=0.98\columnwidth]{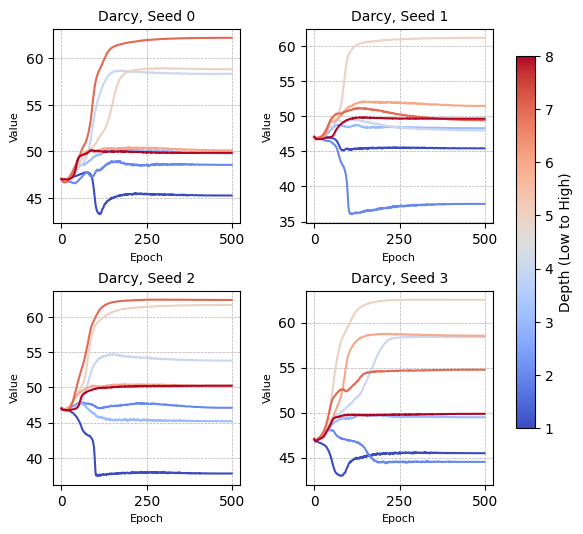} 
\end{center}
\vspace{-0.4cm}
\caption{Parameter evolution across layers for Darcy dataset. 
Visualization of parameter changes in the attention blocks of LA2Former during training on the Darcy dataset across four different random seeds.}
\label{fig:dynamic_k}
\vspace{-0.2cm}
\end{figure}